\documentclass[preprint,12pt]{elsarticle}

\usepackage{amssymb}
\usepackage{hyperref}
\usepackage{makecell}
\usepackage[russian, english]{babel}
\usepackage{amsmath,amsfonts,amssymb}
\usepackage[font=footnotesize,labelfont=bf]{caption}
\usepackage[font=footnotesize,labelfont=bf]{subcaption}

\journal{Journal}

\begin{document}

\begin{frontmatter}



\title{KOHTD: Kazakh Offline Handwritten Text Dataset}


\author[inst1]{Nazgul Toiganbayeva}

\author[inst4]{Mahmoud Kasem}

\author[inst3,inst4]{Galymzhan Abdimanap}

\author[inst3,inst4]{Kairat Bostanbekov}

\author[inst3,inst5]{Abdelrahman Abdallah}

\author[inst3,inst4]{Anel Alimova}

\author[inst3,inst4]{Daniyar Nurseitov}

%
\affiliation[inst1]{
    organization={Al-Farabi Kazakh National University},
    city={Almaty},
    postcode={050040}, 
    country={Kazakhstan}
}

\affiliation[inst3]{
    organization={KazMunayGas Engineering LLP},
    city={Nur-Sultan},
    postcode={010000}, 
    country={Kazakhstan}
}

\affiliation[inst4]{
    organization={Satbayev University},
    city={Almaty},
    postcode={050013}, 
    country={Kazakhstan}
}
\affiliation[inst5]{organization={Information Technology Department },
            addressline={Assiut University}, 
            city={Assiut},
            postcode={71515}, 
            state={Assiut},
            country={Egypt}}
\begin{abstract}
Despite the transition to digital information exchange, many documents, such as invoices, taxes, memos and questionnaires, historical data, and answers to exam questions, still require handwritten inputs. In this regard, there is a need to implement Handwritten Text Recognition (HTR) which is an automatic way to decrypt records using a computer. Handwriting recognition is challenging because of the virtually infinite number of ways a person can write the same message. For this proposal we introduce Kazakh handwritten text recognition research, a comprehensive dataset of Kazakh handwritten texts is necessary. This is particularly true given the lack of a dataset for handwritten Kazakh text. In this paper, we proposed our extensive Kazakh offline Handwritten Text dataset (KOHTD), which has 3000 handwritten exam papers and  more than 140335 segmented images and there are approximately 922010 symbols. It can serve researchers in the field of handwriting recognition tasks by using deep and machine learning. We used a variety of popular text recognition methods for word and line recognition in our studies, including CTC-based and attention-based methods. The findings demonstrate KOHTD's diversity. Also, we proposed a Genetic Algorithm (GA) for line and word segmentation based on random enumeration of a parameter. The dataset and GA code are available at \url{https://github.com/abdoelsayed2016/KOHTD}. 
\end{abstract}


\end{frontmatter}


\section{Introduction}
\label{sec:sample1}
 Deep learning has been widely used in several fields nowadays, such as Medical applications like Cancers diagnoses, detection, and classification \cite{fakoor2013using}, image captioning \cite{huang2019attention},  speech recognition \cite{chorowski2015attention} and in medical question answers \cite{nie2015disease,abdallah2020automated,yu2014deep}, table detection and recognition \cite{abdallah2021tncr,prasad2020cascadetabnet} also deep learning has been used in software engineering such as optimizing the time and schedule of the software projects \cite{waschneck2018optimization,hamada2021neural} and one of the most usages of Deep Learning is handwritten recognition for different languages as we will discuss.
 
 For some types of images, automatic recognition of handwritten texts is still a challenging issue in spite of the recent improvements of the recognition methods and systems, in recent years, handwritten text recognition is attracting more researchers to work on it. Comprehensive and unbounded handwritten datasets are gaining more importance than before, handwritten text can be found: handwritten notes, memos, whiteboards, medical records, historical documents, stylus input text, etc. Therefore, support for understanding the handwritten text in images needs to be provided in a full OCR solution.  For several languages and scripts, this highlights the need for research in the field of developing large-scale handwriting recognition systems. During the last thirty years researchers have been made different types of handwriting text recognition for many languages like English \cite{fischer2013fast,liu2005handwritten,zamora2014neural}, Russian \cite{nurseitov2021handwritten}, Arabic\cite{mahmoud2014khatt,parvez2013arabic}, Malayalam \cite{jomy2013system}, Japanese \cite{das2014algorithm}, etc. 
 
 Any language has a huge number of words. For example, The Oxford dictionary for the English language contains more than 300,000 words. A dictionary for the Kazakh language has more than 380,000 words \cite{sozdikqor}, so it seems impossible to collect a handwritten word database that includes all words. As far as we know,  for the Kazakh language, there is no available public dataset.
 
In our research, we describe a large dataset, called Kazakh Offline Handwritten Text Dataset (KOHTD) to address challenging detection and recognition issues of handwritten Kazakh text in the scanned documents. We present a new Kazakh database for offline handwriting recognition. The dataset is written in Cyrillic and shares the same 42 characters in Kazakh. This dataset is a collection of exam papers from students. There are approximately 922010 symbols in the KOHTD dataset and 140335 segmented images. KOHTD is suggested for many reasons. First, this dataset can serve researchers in the field of handwriting recognition issues by using deep and machine learning.  Second, it’s also a standard and pure dataset for evaluating and comparing different algorithm’s performances. Third, there is no available dataset in the Kazakh language.
 
\begin{figure}[h!]
    \begin{subfigure}{0.50\columnwidth}
        \includegraphics[width=\textwidth]{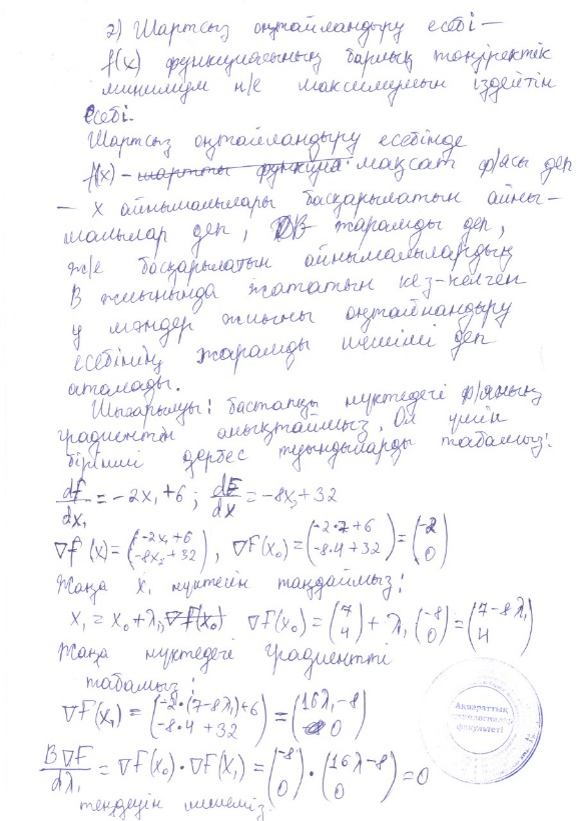}
        \caption{}
    \end{subfigure}
    \begin{subfigure}{0.50\columnwidth}
        \includegraphics[width=\textwidth]{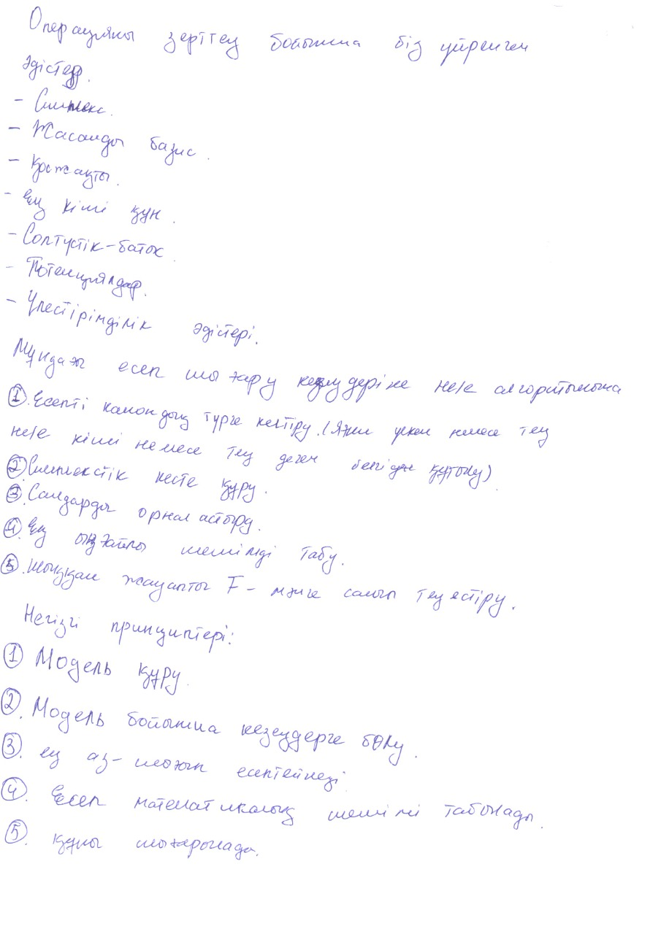}
        \caption{}
    \end{subfigure}
\caption{Examples of images in our dataset} 
\label{fig:examples_images}
\end{figure}

Our database consists of a large collection of exam papers filled by students at Satbayev University and Al-Farabi Kazakh National University, this exam was made and answered in the Kazakh Language (99\%) and Russian Language (1\%) as shown in Fig. \ref{fig:examples_images}, after we received this exam answer, we scanned it and make experiments that related to pre-processing of the examination lists to automatically identifying lists, evaluate the contours of lists, recovering rotations, and also segmentation by line and by words so we can apply our Deep Learning model to recognize each word and remove the artifacts in the edges at the boundaries of segmented words.

We have developed our intelligent software using state-of-the-art deep learning models to solve the problem of recognizing and processing natural language, which consists of optical character recognition of the manuscript texts in Kazakh and Russian languages.

The following section defines the related work on Handwriting Databases and Deep Learning models for handwritten. Section \ref{section:storage} presents the Data collection and storage phases as one of the most time-consuming and costly stages. Section \ref{section:segmentation} provides Dataset segmentation. Section \ref{section:Experimental} provides Experiment Result on the KOHTD dataset and conclusion and future work are given in Section. \ref{section:Conclusion}.

\section{Related Work}
\subsection{Existing Datasets}
IAM dataset \cite{marti1999full,marti2002iam} is a handwritten sentence for the English language. The database can be used for handwritten recognition problems. This database is made on Lancaster-Oslo/Bergen (LOB) Corpus. The IAM Handwriting Database 3.0 is made by 657 different writers and contains 1,539 scanned handwritten pages with 5,685 labeled sentences and 13,353 labeled text lines with a total of 115,320 labeled words, these database has been used in offline handwriting recognition \cite{bunke2004offline,dreuw2011hierarchical,gatos2006hybrid}, handwritten text segmentation \cite{salvi2013handwritten,dos2009text} and writer identification \cite{bensefia2005writer,daniels2013discriminating}.

RAMIS \cite{augustin2006rimes} is a database of an industrial application. The main reason to develop the database was to collect handwritten samples similar to those sent by postal mail and fax by individuals to different companies. The database was filled by 1300 volunteers who contributed to the data collection, providing 5605 mails that contain 12,723 pages. Every mail contains two to three pages, including the letter written by the volunteer, a form of the letter information, and an optimal fax sheet. Then the pages were scanned and the database was published to support testing of the tasks such as mail classification \cite{kermorvant2010handwritten}, handwritten recognition \cite{guichard2010handwritten}, and writer recognition\cite{siddiqi2010text}.

The HKR \cite{nurseitov2021handwritten} is a database of Russian and Kazakh texts that can be used to address detection and recognition problems, the database has 95\% Russian and 5\% Kazakh words/sentences, It’s written in Cyrillic and share 33 characters and there are 9 additional characters for Kazakh alphabet. The dataset is consisting of more than 1,500 forms. The database contains about 63,000 sentences which are more than 715,699 symbols. The HKR database was written by approximately 200 different writers.

The IFN/ENIT \cite{pechwitz2002ifn} is a database of handwritten Arabic town/village names. The forms are filled by 411 writers with nearly 26400 names that contain more than 210000 characters. IFN/ENIT database contains 26459 handwritten Tunisian town/village names. The database is developed for training and determining handwritten Arabic word recognition systems. 

KHATT \cite{mahmoud2014khatt} is a database of Arabic handwritten text, it can be used for Arabic offline handwritten text recognition. KHATT is consisting of 1000 handwritten forms that are written by 1000 different writers. These forms are scanned at 200,300 and 600 dpi resolutions, the database contains 2000 randomly selected paragraphs which consists of 9327 lines. These forms were randomly divided into 70\% for training, 15\% for testing, and 15\% for verification. The database was employed in text recognition, writer identification, and verification.

HIT-MW \cite{su2007corpus} is a database for Chinese handwriting text, it can be used for offline Chinese handwritten text recognition problems. The current version of the HIT-MW database contains 853 forms and 186,444 characters. The database is collected by postal mail and middleman not face to face, it can serve many applications concerning real handwriting recognition.

\subsection{Handwritten Deep Learning Models}
Approaches to handwritten text recognition can be classified into the following categories: Techniques based on HMMs and approaches based on RNNs.
For cursive text recognition, HMM-based approaches have a number of advantages. HMM, models are resistant to noise and can tolerate variations in writing; there are automated algorithms for training the HMM models, and the HMM tools are freely available. Cursive text segmentation is error-prone and time-consuming, which is not required by HMM.

Bunke \cite{bunke2004offline} proposes a system for offline recognition of unconstrained handwritten texts with a wide vocabulary. Only one assumption is made regarding the data: it is written in English. This enables us to apply Statistical Language Models to improve the performance of their system. Data from single and numerous writers have been used in several experiments. Lexica of various sizes (between 10,000 and 50,000 words) were utilized. The usage of language models has been found to improve the system's accuracy. their strategy is detailed in-depth and compared to other ways for dealing with the same problem that has been given in the literature. It is suggested that an experimental configuration be used to correctly deal with unconstrained text recognition.

Safabakhsh \cite{safabakhsh2005nastaaligh} uses a continuous-density variable-duration hidden Markov model, CDVDHMM \cite{chen1995variable} to present a full method for recognizing Farsi Nastaaligh handwritten words. New techniques are used in the preprocessing step to locate and eliminate ascenders, descenders, dots, and other secondary strokes from the original image after binarization, noise reduction, and linked component specification. After that, a new segmentation method based on upper contour analysis and two other processes is used. This algorithm's major purpose is to avoid the problem of under segmentation. The over-segmentation problem can be solved by taking into account variable duration states in the system. The CDVDHMM models the sequence of obtained sub-characters by determining the right-to-left order. In the feature space, eight features are used to represent symbols, including three Fourier descriptors and five structural and discrete characteristics. This feature vector is size and shifts insensitive. Pure characters (without secondary strokes) and some compound forms of characters in the Nastaaligh handwriting style are considered in the model. 

AlKhateeb \cite{alkhateeb2011offline} Using Hidden Markov Models (HMMs), a word-based offline recognition system is proposed. Preprocessing, feature extraction, and classification are the three stages of the approach. The first step is to segment and normalize the words from the input scripts. Then, using a sliding window moving across each mirrored word image, a set of intensity features is collected from each of the split words. Meanwhile, structure-like information such as the number of subwords and diacritical marks are retrieved. Finally, these characteristics are merged into a classification scheme. Intensity features are utilized to train an HMM classifier, and the results are then re-ranked utilizing structure-like features for a higher recognition rate. Extensive trials were conducted utilizing the IFN/ENIT database, which comprises 32,492 handwritten Arabic words.

Otherwise, RNNs, such as the gated recurrent unit (GRU) \cite{chung2014empirical} and the long short-term memory (LSTM) \cite{hochreiter1997long} can fix this problem. Speech recognition \cite{hannun2014deep}, machine translation \cite{sutskever2014sequence}, video summarising \cite{srivastava2015unsupervised}, and others. sequence-to-sequence learning tasks have demonstrated RNN models' amazing skills. It is necessary to convert a two-dimensional image to a vector and send it to an encoder and decoder in order to transform it for offline HTR. 

GRU and LSTM handle the problem by combining information and features from many sources. RNN networks are fed these handwriting sequences. The input feature does not require segmentation due to the usage of Connectionist Temporal Classification (CTC) \cite{graves2006connectionist} models. One of the main advantages of the CTC algorithm is that it does not require any segmented labeled data. We can employ data alignment with the output.

RR Ingle \cite{ingle2019scalable} focuses on three issues that arise while creating such systems: data, efficiency, and integration. For starters, acquiring large amounts of high-quality training data is one of the most difficult tasks. They solve the challenge by analyzing online handwriting data gathered for a large-scale online handwriting recognition system. They present our picture data generating pipeline and investigate how online data may be used to construct HTR models. They show that when only a few real photos are available, as is frequently the case with HTR models, the data improves the models dramatically. It allows supporting a new script for a much-reduced price. Second, they propose a neural network-based line recognition model without recurrent connections. The model reaches a level of accuracy comparable to LSTM-based models while allowing for more simultaneous training and inference. Finally, they show how to integrate HTR models into an OCR system in a straightforward manner. These components make up a solution for integrating HTR into a large-scale OCR system.

Espana-Boquera \cite{espana2010improving} proposes hybrid Hidden Markov Model (HMM) and Artificial Neural Network (ANN) models for identifying unconstrained offline handwritten texts. Markov chains were employed to describe the structural elements of the optical models, and a Multilayer Perceptron was used to estimate the emission probability. With supervised learning approaches, this work also introduces novel strategies for removing slope and slant from handwritten text and normalizing the size of text images. Slope correction and size normalization are performed by using Multilayer Perceptrons to classify the local extrema of text contours. Artificial Neural Networks are also used to reduce slant in a nonuniform manner. Experiments were conducted using offline handwritten text lines from the IAM database, and the recognition rates attained were among the best for the identical job when compared to those published in the literature.

F Abdurahman \cite{abdurahman2021ahwr}  proposes an offline handwritten Amharic (the language of the Federal Government of Ethiopia) word recognition system based on convolutional recurrent neural networks. Convolutional neural networks (CNNs) for feature extraction from input word images, recurrent neural networks (RNNs) for sequence encoding, and connectionist temporal classification as a loss function are all part of the proposed system. they have created a dataset of handwritten Amharic words, HARD-I. their best-performing recognition model achieved a WER of 5.24 percent and a CER of 1.15 percent from testing on various recognition models utilizing their dataset. When compared to existing models for offline handwritten Amharic word recognition, the proposed models perform well.

\section{Data collection and storage}
\label{section:storage}
The KOHTD database discusses the problem of recognizing manuscripts in Kazakh and Russian languages in relation to Cyrillic graphics, describes and investigates various approaches, as well as presents the results of the study, and suggests the Bluche and Puigcerver models \cite{bluche2017gated, puigcerver2017multidimensional} for comparing the results. The recognition of handwritten text of the Kazakh-Russian language remains not fully investigated. In this regard, the development and research of new effective algorithms for recognizing handwritten text of the Kazakh-Russian language are relevant.
The approach to the problem of handwriting recognition of the Kazakh-Russian language, based on the use of neural networks, is proposed. The main stage of handwriting data collection of the Kazakh-Russian language consists of the following stages:
\begin{itemize}

    \item  Pre-processing for handwriting recognition: at this stage, the image is processed to improve its quality and bring it to a form that is convenient for segmentation. At the pre-processing stage, the handwritten text is scanned with a Canon MF4400 Series UFRII scanner. The resolution for the scanned examination lists is 300 dpi and the color depth is 24 bits. Translation of paper documents into a digital graphical representation. 
    \item  Segmentation of the scanned handwritten text into words: at this stage, the scanned handwritten text is divided, or segmented, into convenient parts for analysis. The most natural actions at this stage are to split the text into separate lines (line segmentation) and split the lines into words (word segmentation), where space is their separator.  To do this, filters are consistently applied to the text to remove noise and determine the boundaries of words.
    
    \item  Annotation of segmented words, which will map each image to its text in the JSON format file.
\end{itemize}

\subsection{Labeling in the database}

The main idea was as follows:
images of segmented words were sent to volunteer users via a Telegram bot named @collectorOfdataset\_bot (botCollector). It is a messenger application for mobile phones that allows you to create programs in Python and register them as bots \cite{telegram}. To simplify the process of collecting annotations, it was decided to send the same image to two random users. In case the sent annotations from both users
turned out to be absolutely identical, we considered this result to be reliable.

After sending out half of the images from the total number and receiving the results, it turned out that most of the users filled out annotations very inattentively, and sometimes even wrote deliberately inappropriate annotations.
To exclude such cases, we trained a neural network for handwriting recognition on an already existing incomplete dataset. Further, after each input and sending of the annotation, we calculated the Levenshtein distance between the result from the user and the result of recognition. To our surprise, even on a partial dataset, the neural network produced subjectively excellent results.
Also, by typing the command /my\_annotations, each user could see their statistics: the number of annotated words, the number of words whose annotations do not match the annotation from another user, the average confidence value.
\begin{figure}[!ht]
    \centering
    \includegraphics[width=0.50\linewidth,height=0.50\textheight,keepaspectratio]{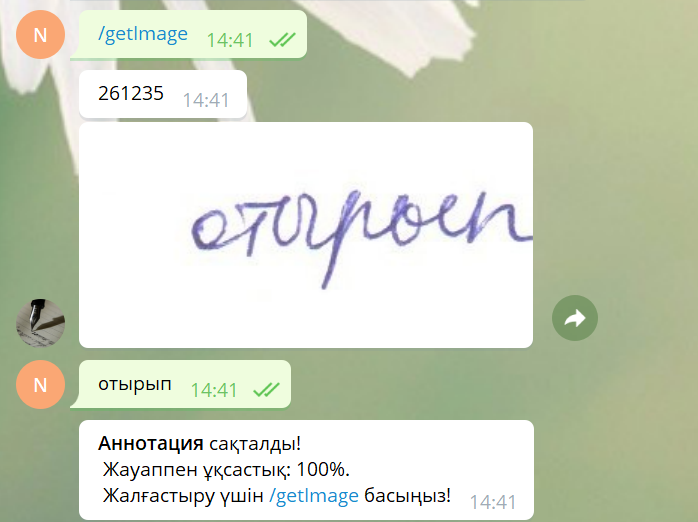}
    \caption{result of the getImage command with 100\% similarity}
    \label{fig:bot_capture}
\end{figure}

To get another picture for annotation, users sent the command /getImage (Figure \ref{fig:bot_capture}).
If the image contains incomprehensible words, only numbers, corrections, words in a foreign language, several lines, the inscription had to be removed by clicking the "Delete" button.
The complexity of the handwriting recognition task is a large variety of handwriting, shapes, sizes of letters, and a variety of languages. Also, the paper with the text may contain ``noises'' – paper defects, foreign spots, which also complicates the whole process.  Handwritten texts, as you know, differ from printed ones in the manifestations of individual handwriting properties: from the calligraphically printed letters of each word in the text to the illegibly written text as a whole, despite the fact that there is a standard for writing a particular letter, and this standard writing was taught in elementary school by every writer. But in the process of practicing writing, each writer develops individual features of handwriting.
For handwriting recognition in the Kazakh language, more than 3 thousand pages in A4 format were scanned from various handwriting samples (handwritten exam answers to students' questions). Table \ref{tab:similarity} shows the recognition of the handwritten text of the Kazakh language of one voluntary user.

At the time of this writing, 643 people were involved in this work (mainly students and university teachers).

\begin{table}[h!]
    \centering
    \caption{Recognition of handwritten text of the Kazakh language.}
    \label{tab:similarity}
    \begin{tabular}{c|c|c|c}
         N & input & Words &Similarity  \\ \hline
         1 &   
            \begin{minipage}{.3\textwidth}
              \includegraphics[width=\linewidth, height=4mm]{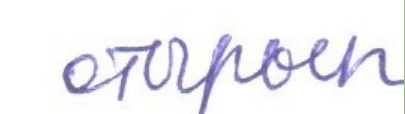}
            \end{minipage}
            & \foreignlanguage{russian}{отырып} & 100\% \\ \hline
         2 & \begin{minipage}{.3\textwidth}
              \includegraphics[width=\linewidth, height=4mm]{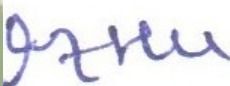}
            \end{minipage}& \foreignlanguage{russian}{яғни} & 75\% \\ \hline
         3 & \begin{minipage}{.3\textwidth}
              \includegraphics[width=\linewidth, height=4mm]{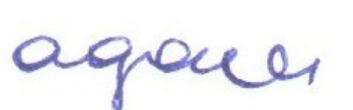}
            \end{minipage}& \foreignlanguage{russian}{адам} & 67\% \\ \hline
         4 & \begin{minipage}{.3\textwidth}
              \includegraphics[width=\linewidth, height=4mm]{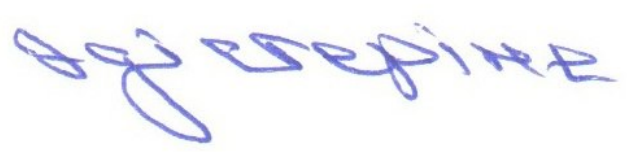}
            \end{minipage}& \foreignlanguage{russian}{әдістеріне} & 84\% \\ \hline
         5 & \begin{minipage}{.3\textwidth}
              \includegraphics[width=\linewidth, height=4mm]{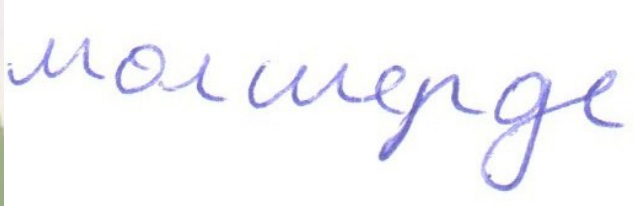}
            \end{minipage}& \foreignlanguage{russian}{мөлшерде} & 100\% \\ \hline
         6 & \begin{minipage}{.3\textwidth}
              \includegraphics[width=\linewidth, height=4mm]{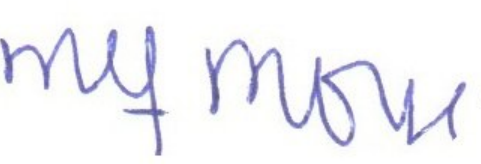}
            \end{minipage}& \foreignlanguage{russian}{тұтын} & 83\% \\ \hline
         7 & \begin{minipage}{.3\textwidth}
              \includegraphics[width=\linewidth, height=4mm]{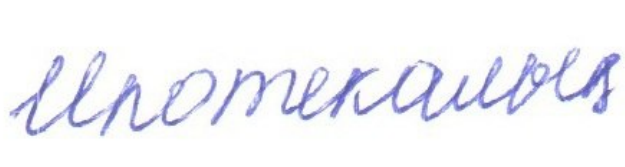}
            \end{minipage}& \foreignlanguage{russian}{Ипотекалық} & 90\% \\ \hline
         8 & \begin{minipage}{.3\textwidth}
              \includegraphics[width=\linewidth, height=4mm]{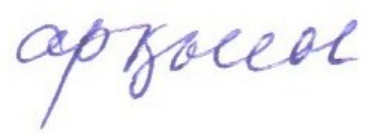}
            \end{minipage}& \foreignlanguage{russian}{арқылы} & 83\% \\ \hline
         9 & \begin{minipage}{.3\textwidth}
              \includegraphics[width=\linewidth, height=4mm]{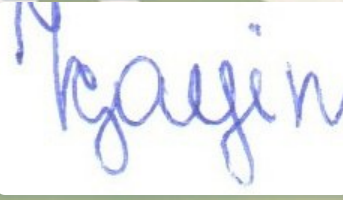}
            \end{minipage}& \foreignlanguage{russian}{қауіп} & 60\% \\ \hline
    \end{tabular}
\end{table}

The table shows that the similarity to the answer depends on the person's handwriting. In the Kazakh language, the letters \foreignlanguage{russian}{``ң'' and ``қ''} (Table \ref{tab:similarity}, Line 7), \foreignlanguage{russian}{``м'' and ``ш''}(Table \ref{tab:similarity}, Line 3), \foreignlanguage{russian}{``н'' and ``и''}(Table \ref{tab:similarity}, Line 6), \foreignlanguage{russian}{``л'' and ``е''}(Table \ref{tab:similarity}, Line 8), are similar in handwritten text. In handwritten texts, the recognition process is complicated by the individual features of the handwriting, including the variability of writing letters. The presence of gaps in the text also contributes to the perception and recognition of the manuscript. They help you not to read the text sequentially from beginning to end. In addition to spaces, capital letters, as well as lowercase letters that come out of an even row of lines, are essential for the perception and recognition of what is written. A voluntary user sometimes enters a word with a capital letter with a lowercase one, which reduces the percentage of similarity with the answer (Table \ref{tab:similarity}, line 9).

\subsection{Characteristics of the Database}
The database is composed of segmented words from over 3000 scanned exam papers. Generally speaking, a separate study needs to be done to estimate the number of people who filled out these forms. But we assume that on average one person completed 2 examination sheets. Accordingly, according to our estimates, we have approximately 1000 different styles of handwriting paper.
By the number of people who annotated the images, we can say for sure that at the time of writing this article, there were 643 of them, since we have all the records in the database.

\begin{figure}[h!]
    \centering
    \includegraphics[width=0.5\linewidth,height=0.5\textheight,keepaspectratio]{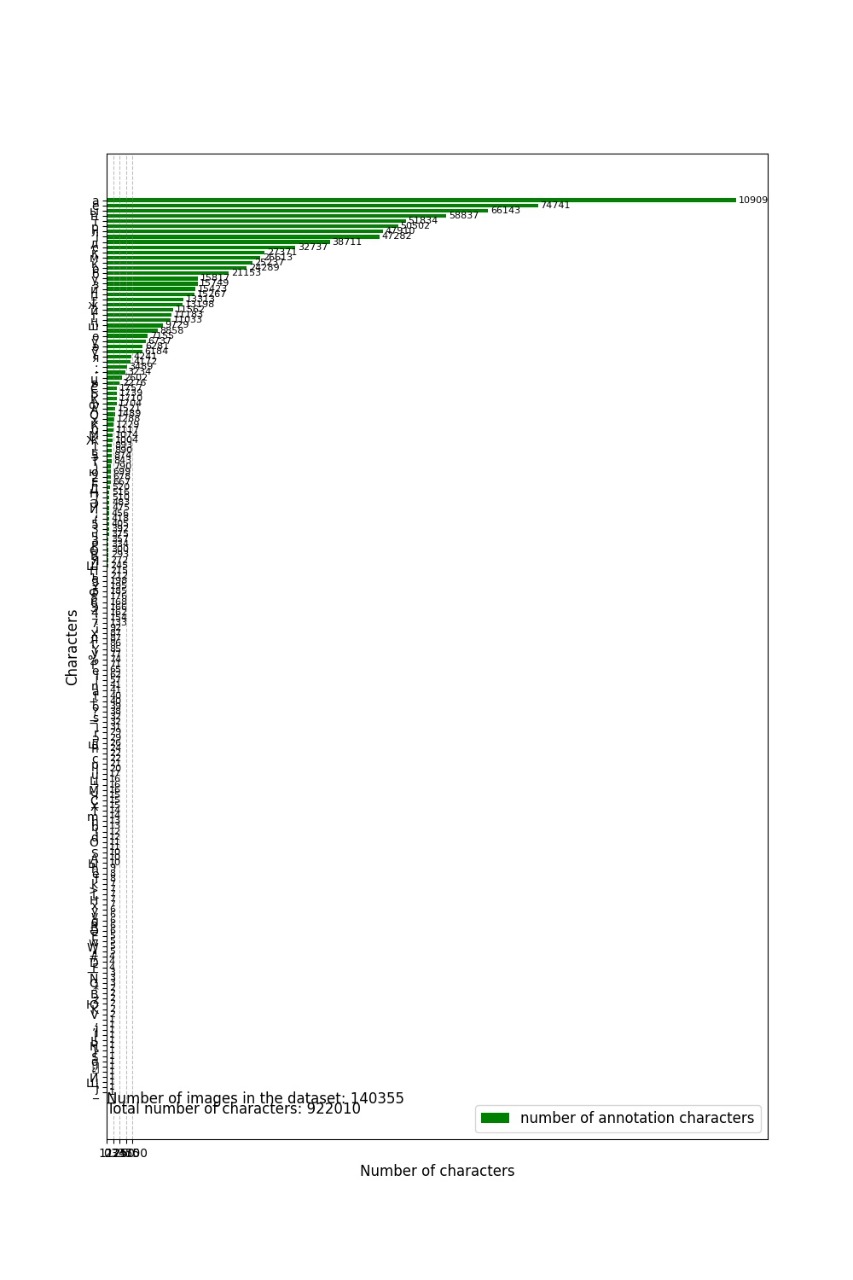}
    \caption{Histogram of Characters in the dataset}
    \label{fig:dataset-histogram}
\end{figure}

There are approximately 922010 symbols shown in Figure \ref{fig:dataset-histogram}. The total number of images in the dataset is 140335 images after pre-processing and segmentation of the examination lists.

\subsection{Statistic Analysis }
In this section, statistics on the database are presented. The database is split into three exclusive parts( training, validation, and test sets). Table \ref{tab:table_staticstic} shows the uni-, bi-, and tri-grams of these sets and of the full database and also shows how many words, unique words, and a number of characters for all the dataset and the parts.

\begin{table}[h!]
    \centering
    \caption{N-gram statistics of the database}
    \label{tab:table_staticstic}
    \begin{tabular}{|c|c|c|c|c|c|c|}
   \hline
         Set &  \makecell{Word \\ count}   &  \makecell{Unique \\ word}    & \makecell{Character \\ count}     & \makecell{Uni-\\grams} & \makecell{Bi-\\grams} & \makecell{Tri-\\grams} \\ \hline
         Training   & 104278 &  24943 &   98258  & 24943 &   100142  &  104242      \\ \hline
         Validation & 22386  &  8870  &   21054  & 8870  &   22076   &  22376    \\ \hline
         Testing    & 22351  &  8984  &   21054  & 21054 &   22040   &  22346     \\ \hline
         All data   & 149015 & 31483  &  140366  & 31483 &   130912  &   147238  \\ \hline
    \end{tabular}
\end{table}

Table \ref{tab:oov} shows the out of vocabulary (OOV) statistics of the validation and test sets compared with the training data sets.
\begin{table}[h!]
    \centering
    \caption{N-gram statistics of the database}
    \label{tab:oov}
    \begin{tabular}{|c|c|c|}
   \hline
         Out of Vocabulary &  OOV Tokens & OOV Percentage \\ \hline
         Validation &  3408 & 10.36\% \\ \hline
         Testing &     3442 & 10.46\% \\ \hline
    \end{tabular}
\end{table}

\section{Dataset segmentation}
\label{section:segmentation}
Segmentation of text into lines is one of the most important steps in the optical character recognition (OCR) process, in particular, in the optical recognition of document images. Line segmentation is the decomposition of an image containing a sequence of characters into fragments containing individual characters.

The importance of segmentation is due to the fact that the majority of modern OCR systems are based on classifiers (including neural networks) of individual characters, and not words or text fragments. In such systems, errors of incorrect insertion of cuts between characters, as a rule, are the cause of the lion's share of errors in the final recognition.

In our case, for line segmentation, we used a genetic algorithm. Genetic Algorithm (GA) is a classic evolutionary algorithm based on random enumeration of a parameter. By random here we mean that in order to find a solution using GA, random changes were applied to the current solutions to generate new ones. GA is based on Darwin's theory of evolution. It is a slow, gradual process that works by making small and slow changes. In addition, GA is slowly making small changes to its decisions until it gets the best solution.

\begin{figure}[!ht]
    \centering
    \includegraphics[width=0.75\linewidth,height=0.75\textheight,keepaspectratio]{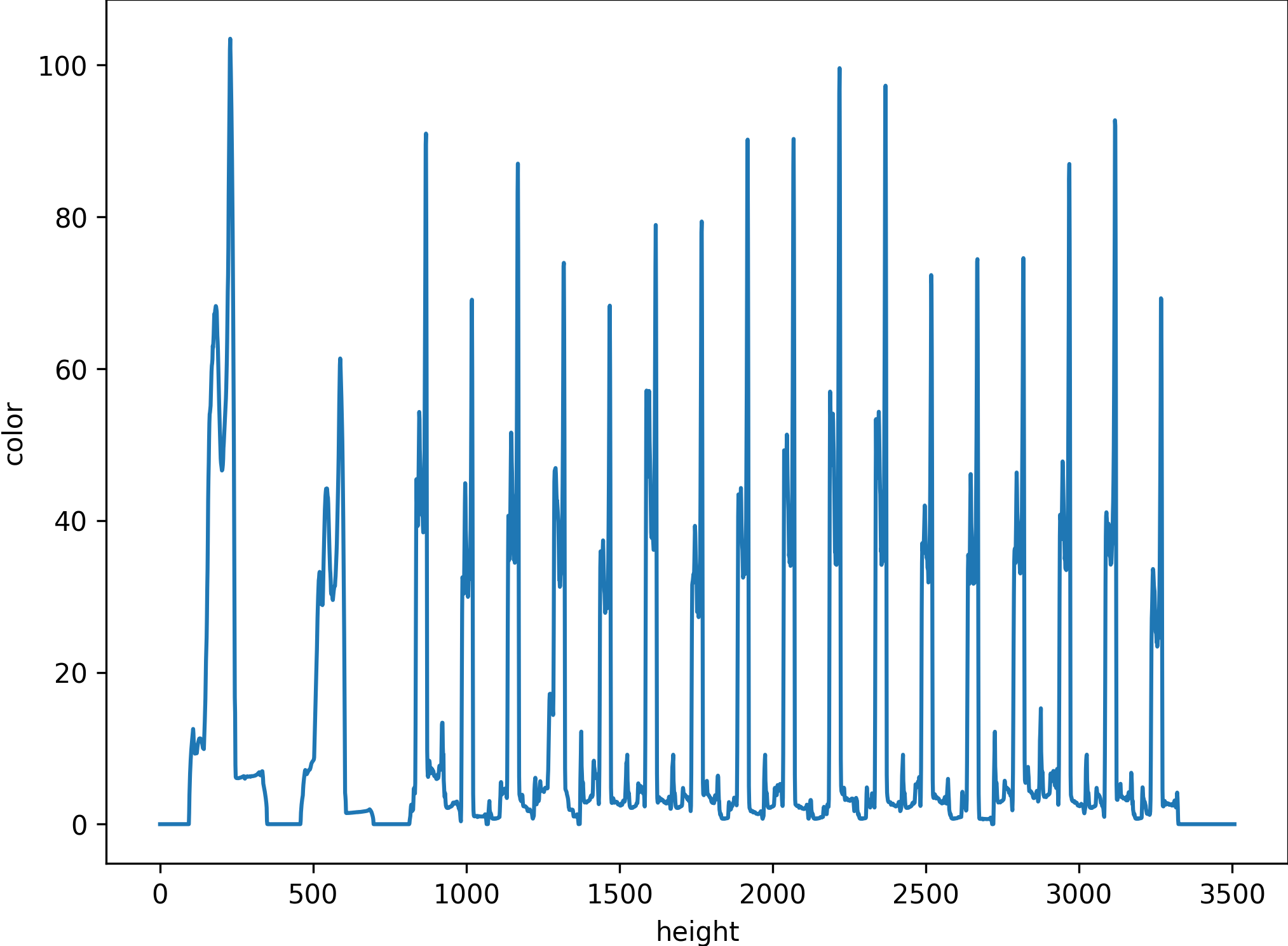}
    \caption{Image histogram by height}
    \label{fig:segm1}
\end{figure}

To implement the genetic algorithm, first of all, we need to determine the coordinates from where we will start making changes. For correct segmentation, the coordinates must lie between two lines. To determine the extreme points, we get a histogram of the picture in height.
If the line is completely white, then the sum of pixels will be equal to 255, for convenience we will make an inversion, then the white color will be equal to zero. So our optimization function should tend to zero. The result is shown in Figure \ref{fig:segm1}.

\begin{figure}[!ht]
    \centering
    \includegraphics[width=0.75\linewidth,height=0.75\textheight,keepaspectratio]{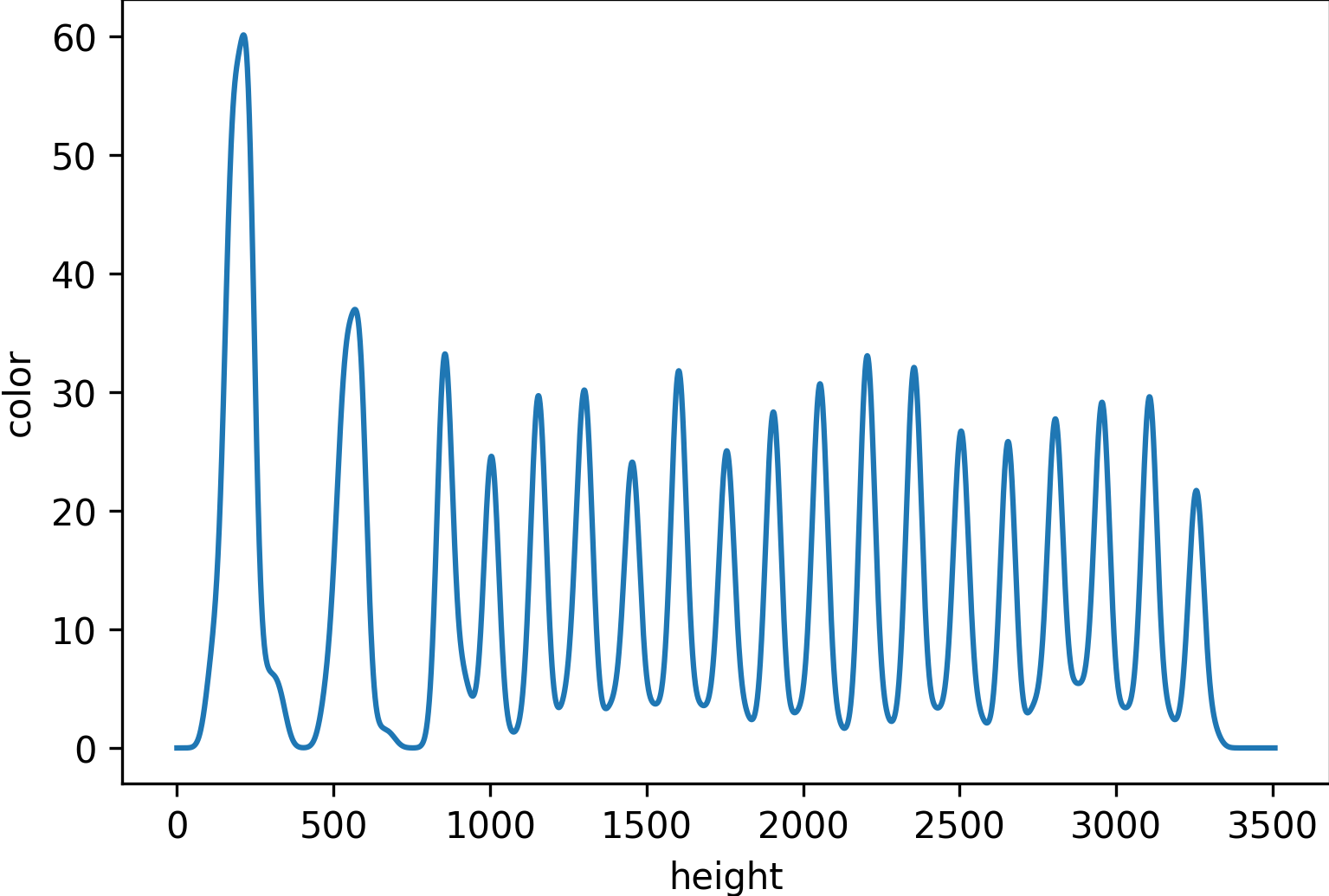}
    \caption{Smoothed image histogram by height}
    \label{fig:segm2}
\end{figure}

\begin{figure}[!ht]
    \centering
    \includegraphics[width=0.75\linewidth,height=0.75\textheight,keepaspectratio]{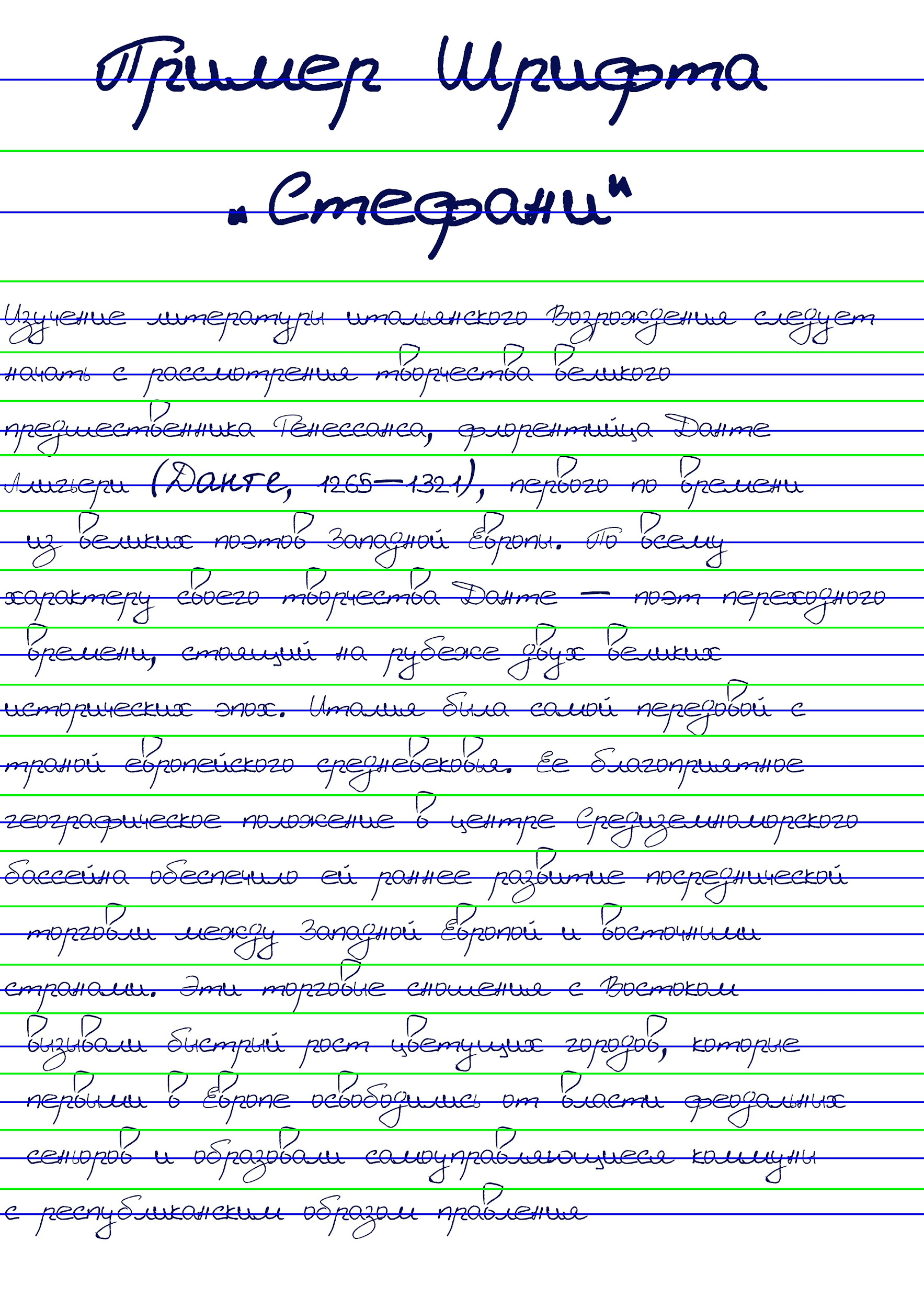}
    \caption{Line drawing}
    \label{fig:segm3}
\end{figure}

The histogram gives us the sum of the pixels of each line of the image, to get explicit vertices, we smooth the data using a gaussian filter. The result is shown in Figure \ref{fig:segm2}.
Let's draw lines at the maximum (blue) and minimum (green) points on the histogram, we get Figure \ref{fig:segm3}.

\begin{figure}[!ht]
    \centering
    \includegraphics[width=0.75\linewidth,height=0.75\textheight,keepaspectratio]{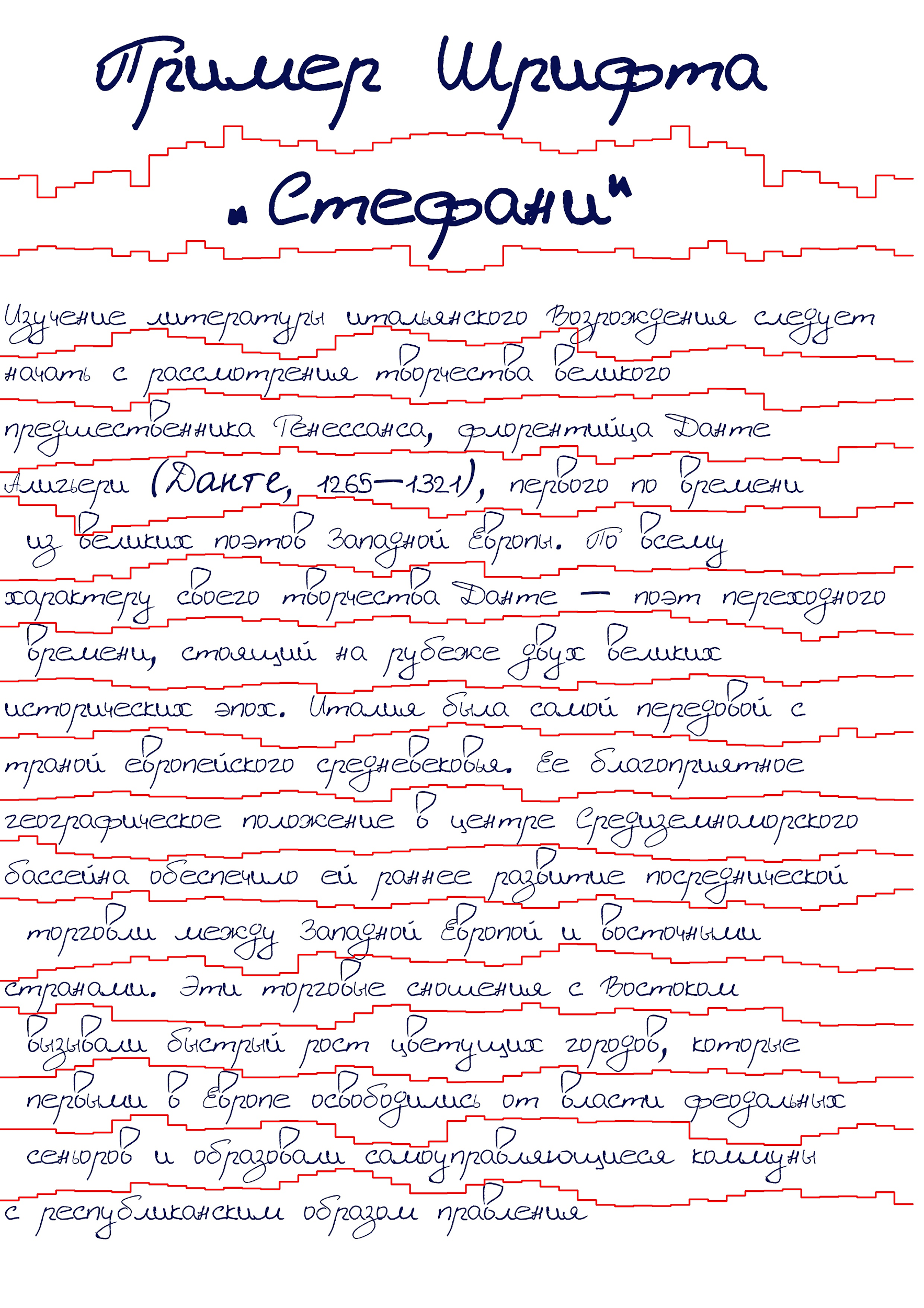}
    \caption{Genetic Algorithm Result, String Segmentation}
    \label{fig:segm4}
\end{figure}

Our coordinates will lie between the blue lines. After that, we run the genetic algorithm and get the line segmentation. The result is in Figure \ref{fig:segm4}.

After receiving the lines, we need to segment the words, since our model accepts words as input. We will do the segmentation of words in the same way, through histograms. First, draw a histogram horizontally and smooth it with a Gaussian filter. Find the vertices and draw in the picture. We get the result as in Figure \ref{fig:segm4}.
\begin{figure}[!ht]
    \centering
    \includegraphics[width=0.75\linewidth,height=0.75\textheight,keepaspectratio]{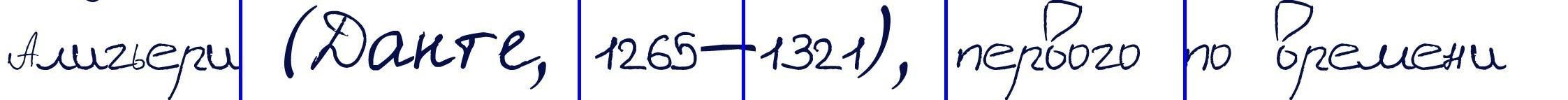}
    \caption{Segmentation of words}
    \label{fig:segm5}
\end{figure}

\section{Experimental results}
\label{section:Experimental}

A quantitative comparison of well-known recurrent neural networks (RNN), such as Bluche\cite{bluche2017gated}, Puigcerver\cite{puigcerver2017multidimensional}, Flor \cite{de2020htr} and Abdallah\cite{Abdallah_2020} models, has been implemented to choose the best performing model on the dataset given. At the first, the final dataset was split into three datasets as follows: Training (70\%), Validation (15\%), and Testing (15\%).  After training, validation, and testing datasets were prepared, the models were trained, and a series of comparative evaluation experiments were conducted. As experiment results proved, the Flor model demonstrated the best performance with a 6.52 \% character error rate (CER), 24.52\% word error rate (WER), and 26.98\% SER  for the test dataset.

\subsection{Evaluation methods}
We used two ways to evaluate models in this article: For the results reported in the first technique, established performance measurements such as the character error rate (CER) and word error rate (WER)\cite{frinken2014continuous} are employed. The Levenshtein distance is calculated by dividing the total number of characters in the ground truth word ($N$) by the amount of character substitution ($S$), insertion ($I$), and deletions ($D$) required to convert one string into another.

\begin{equation}
\begin{split}
CER = \frac{S+I+D}{N}	
\end{split}
\end{equation}
Similarly, the $WER$ is determined by the sum of the number of term substitutions ($S w$), insertions ($I w$), and deletions ($D w$) required for the transformation of one string into another, and divided by the total number of ground-truth terms ($N w$). 

\begin{equation}
\begin{split}
WER = \frac{S_w + I_w + D_w}{N_w}	
\end{split}
\end{equation}

\subsection{Training}
Tensorflow\cite{abadi2016tensorflow}, a Python-based deep learning library, was used to train all of the models. Through Python, Tensorflow allows for the transparent usage of highly efficient mathematical operations on GPUs. In the Python script, a computational graph is built to define all operations required for the given computations.

The machine that was carried out the tests with 2x Intel(R) Xeon(R) E-5-2680 processors, 4x NVIDIA Tesla k20x graphics cards, and 100 GB of RAM. The usage of a GPU reduced model training time by about a factor of three, but this speed-up was not extensively evaluated during the project, so it may have varied.

The report's plots were built with the Python package matplotlib, and the visuals were developed with Inkscape, a vector graphics program akin to Adobe Photoshop.

The validation loss value is minimized in all models. The RMSProp method\cite{hinton2012neural} is used to execute the stochastic gradient descent optimization using a base learning rate of 0.001 and 32 mini-batches. We also used early stopping with patience 20 since we wanted to track the validation loss at each epoch and terminate training if the validation loss did not improve after 20 epochs.

\subsection{Proposed Models}
In this section, we will present deep learning models based on RNN and CTC loss function which has been implemented to choose the best performing model on the dataset given. Bluche \cite{bluche2017gated} built a deep neural network that could split into three main parts, First convolutional layers as the encoder of the input images. It operates two-dimensional representations and supplies 2D features maps. This part contains about 20\% of the model’s free parameters but represents the slowest component of their architecture. Aims to make it generic to be reusable to factorize the processing time. The second is the interface that transforms the 2D image-like representation into 1D representation. The third is the decoder, a bidirectional Long Short Term Memory Recurrent Neural Network (LSTM RNN) that processes feature sequences to predict a sequence of characters. This part has the most capacity of the network about 80\% but has fast processing.

Puigcerver \cite{puigcerver2017multidimensional} present a neural network architecture based on convolutional and 1D-LSTM layers to make line-level Handwritten Text Recognition (HTR). Then providing statistically sound empirical study that proves the architecture gives similar or better accuracy compared with the state of the art 2D-LSTM architecture and incomparably faster. And showing that performing appropriate random distortions on the training images, reduce the error rates. 

Flor\cite{de2020htr} proposed New Gated CRNN architecture used for offline Handwritten Text Recognition (HTR) systems, by using the latest machine learning techniques and approaches in the field of Natural Language Processing (NLP), like the Gated mechanism presented by Dauphin \cite{dauphin2017language}, and Bidirectional Gated Recurrent Units (BGRU) \cite{cho2014learning}, Flor proposed Gated-CNN-BGRU model which involves a few parameters and achieves a low error rate in the text recognition process. There are more contributions with the following aspects, handling long sentences with different styles, noise, and variations, even in the case of limited training data. Improving the recognition results through the new Gated-CNN-BGRU architecture. Reducing the number of trainable parameters, making the model smaller and with lower computational cost.

Abdallah \cite{Abdallah_2020}, aiming at improving HTR model accuracy in the handwritten Cyrillic text recognition task. This model's architecture consists of 4 main parts: encoder, attention, decoder, and CTC.
An encoder part consists of 5 convolutional blocks, each of which is made up of a convolutional layer, Parametric Rectified Linear Unit (PReLU) activator \cite{he2015delving} with Batch Normalization, and gated convolutional layer \cite{bluche2017gated}. The Dropout technique is also applied at the input of some convolutional layers (with a dropout probability of 0.5) to reduce the overfitting issue \cite{srivastava2014dropout}.
As an attention part of this model's architecture, the Bahdanau attention mechanism is used \cite{bahdanau2015neural}. Generally, attention mechanisms encode an input sentence by segmenting it into a fixed number of parts so they can be processed later by a decoder. Bahdanau's attention mechanism enabled the attention mechanism to focus on relevant parts of an input sentence, rather than hard segmenting it. The key role of the Bahdanau attention mechanism applied between an encoder and decoder is to provide a richer encoding of the input sequence.

\subsection{Results}

All the models were trained on the KOHTD dataset. We evaluated these models by the standard performance measures used for all results presented: CER, WER, and SER. For all models, the minibatch of 32 size and Early Stopping after 20 epochs without improvement in validation loss value and lr=0.001 were set. For the best use of each model, within the 20 tolerance epochs, ReduceLRonPlateau schedule\cite{vinciarelli2001new} with a decay factor of 0.2 after 10 epochs without improvement in validation loss value was also used. All of the following figures present the character error rate, which shows how the model detects each character.

\begin{figure}[h!]
    \begin{subfigure}{0.50\columnwidth}
        \includegraphics[width=\textwidth]{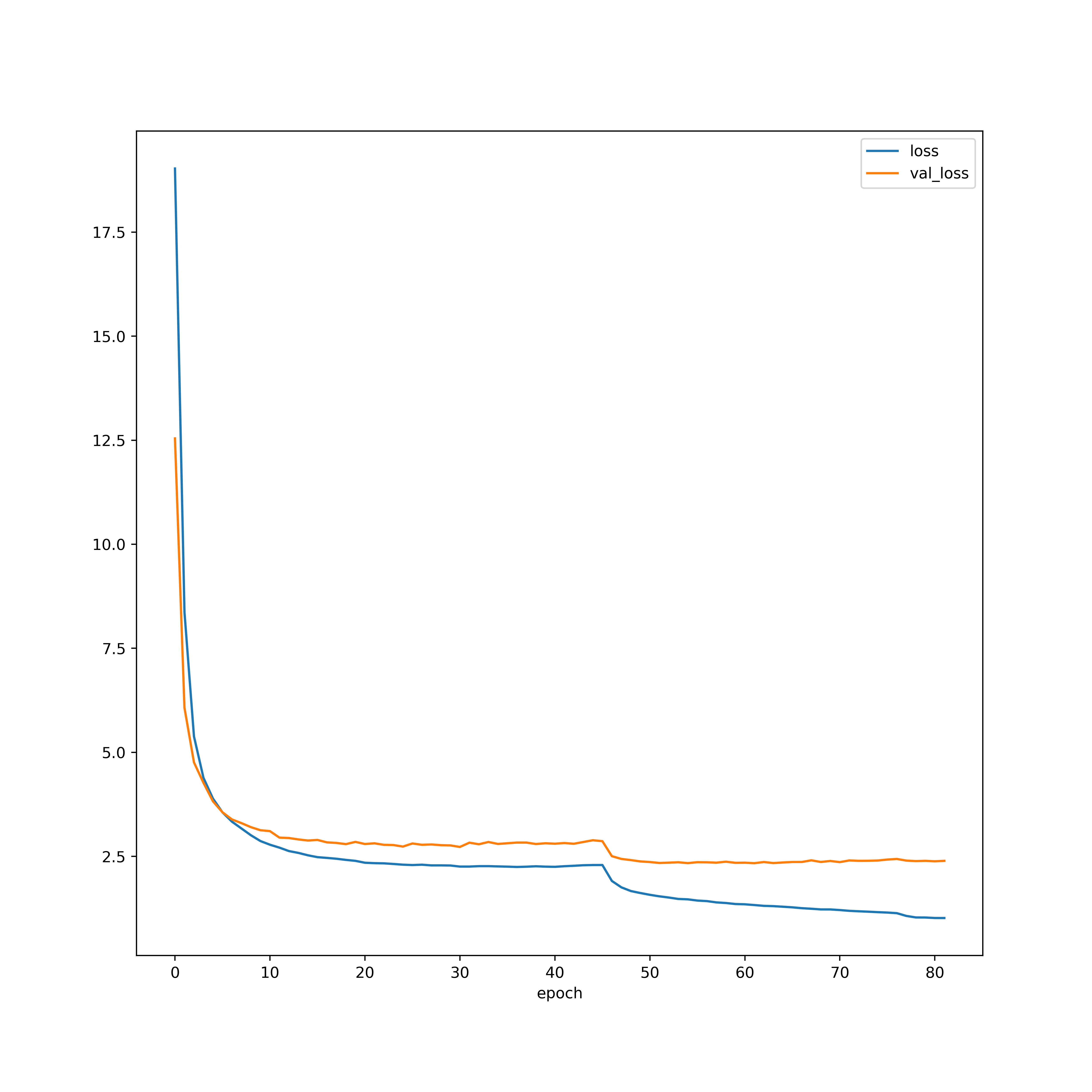}
        \caption{Bluche model loss}
        \label{fig:bluche_loss}
    \end{subfigure}
    \begin{subfigure}{0.50\columnwidth}
        \includegraphics[width=\textwidth]{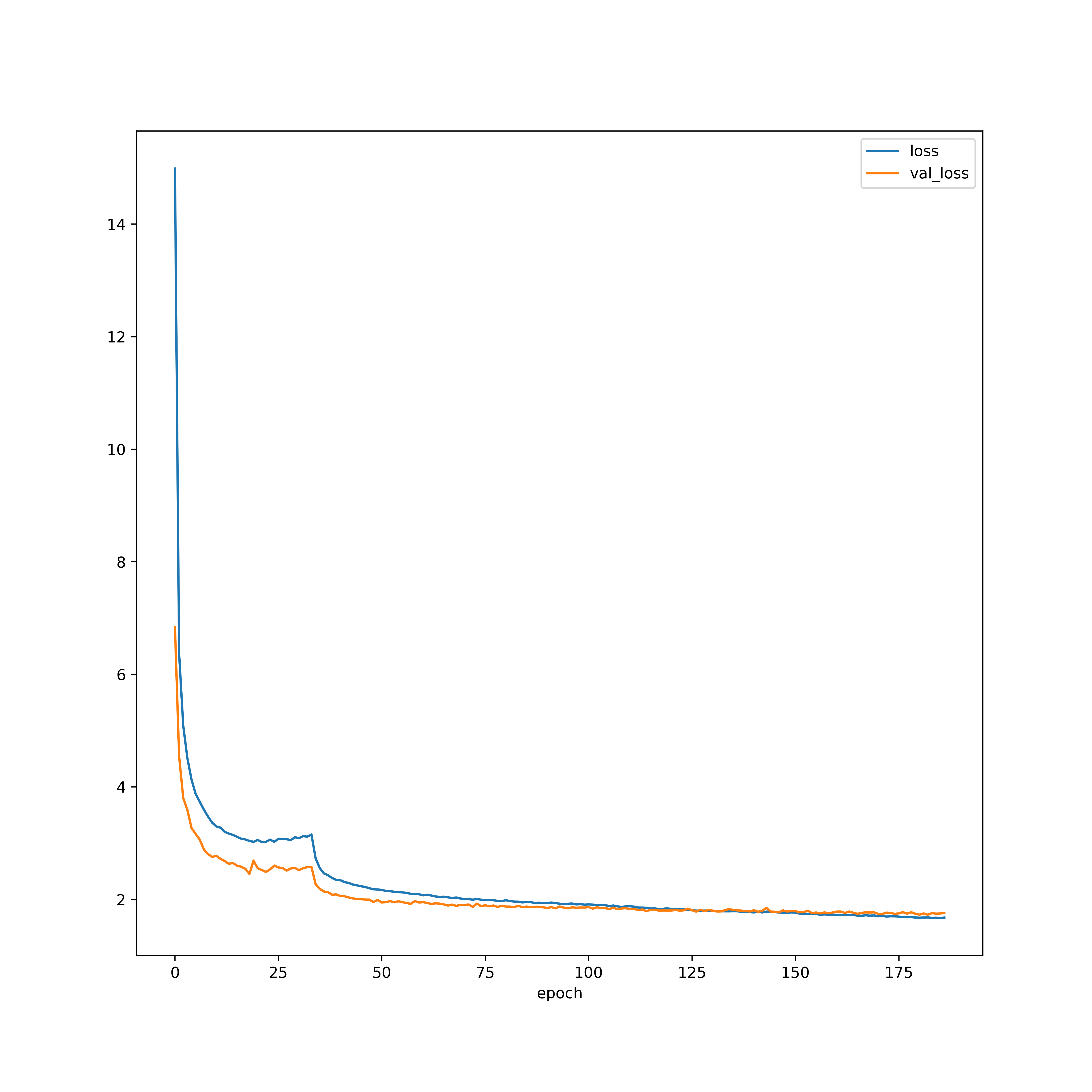}
        \caption{Flor model loss}
        \label{fig:flor_loss}
    \end{subfigure}
    \begin{subfigure}{0.50\columnwidth}
        \includegraphics[width=\textwidth]{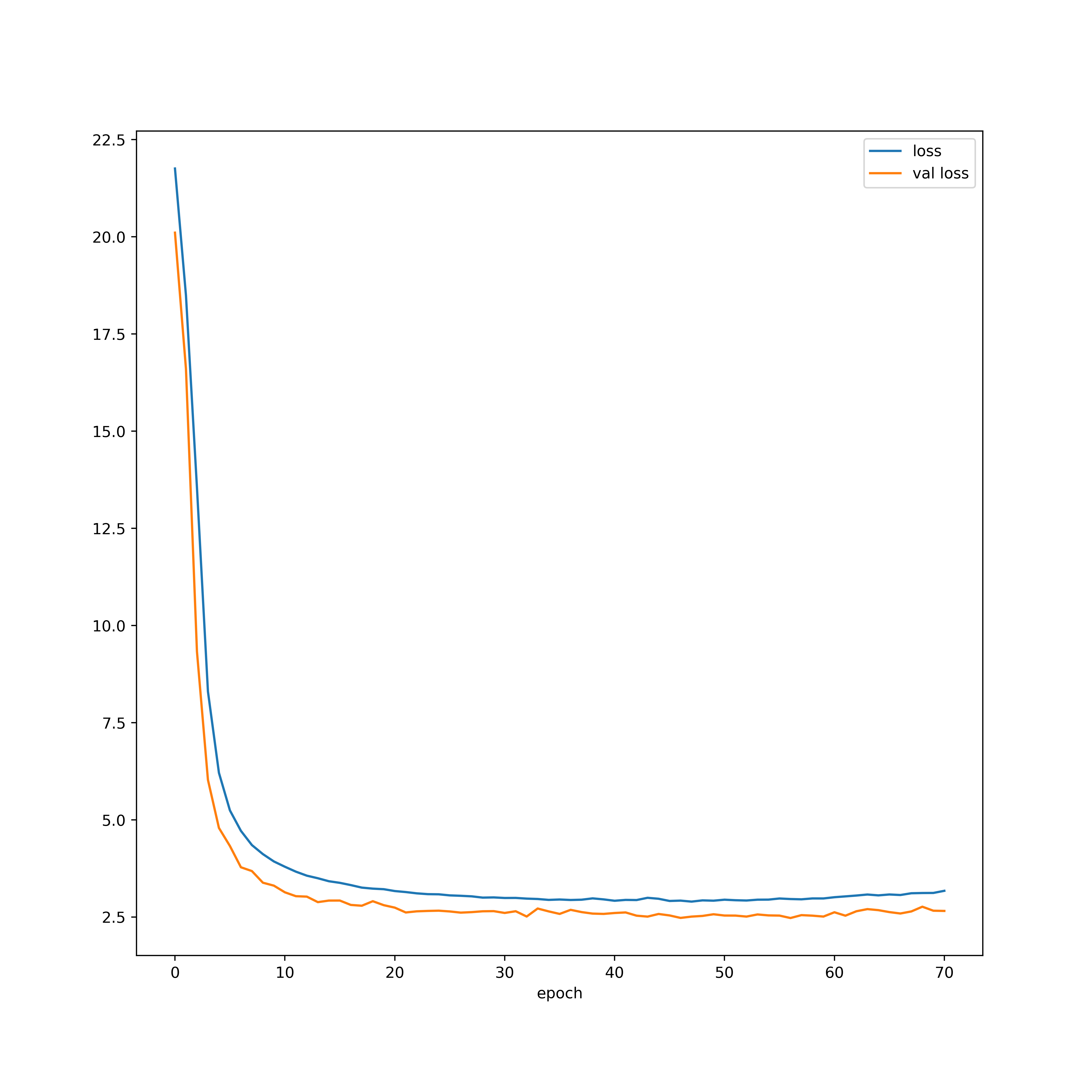}
        \caption{Abdallah model loss}
        \label{fig:abdo_loss}
    \end{subfigure}
    \begin{subfigure}{0.50\columnwidth}
        \includegraphics[width=\textwidth]{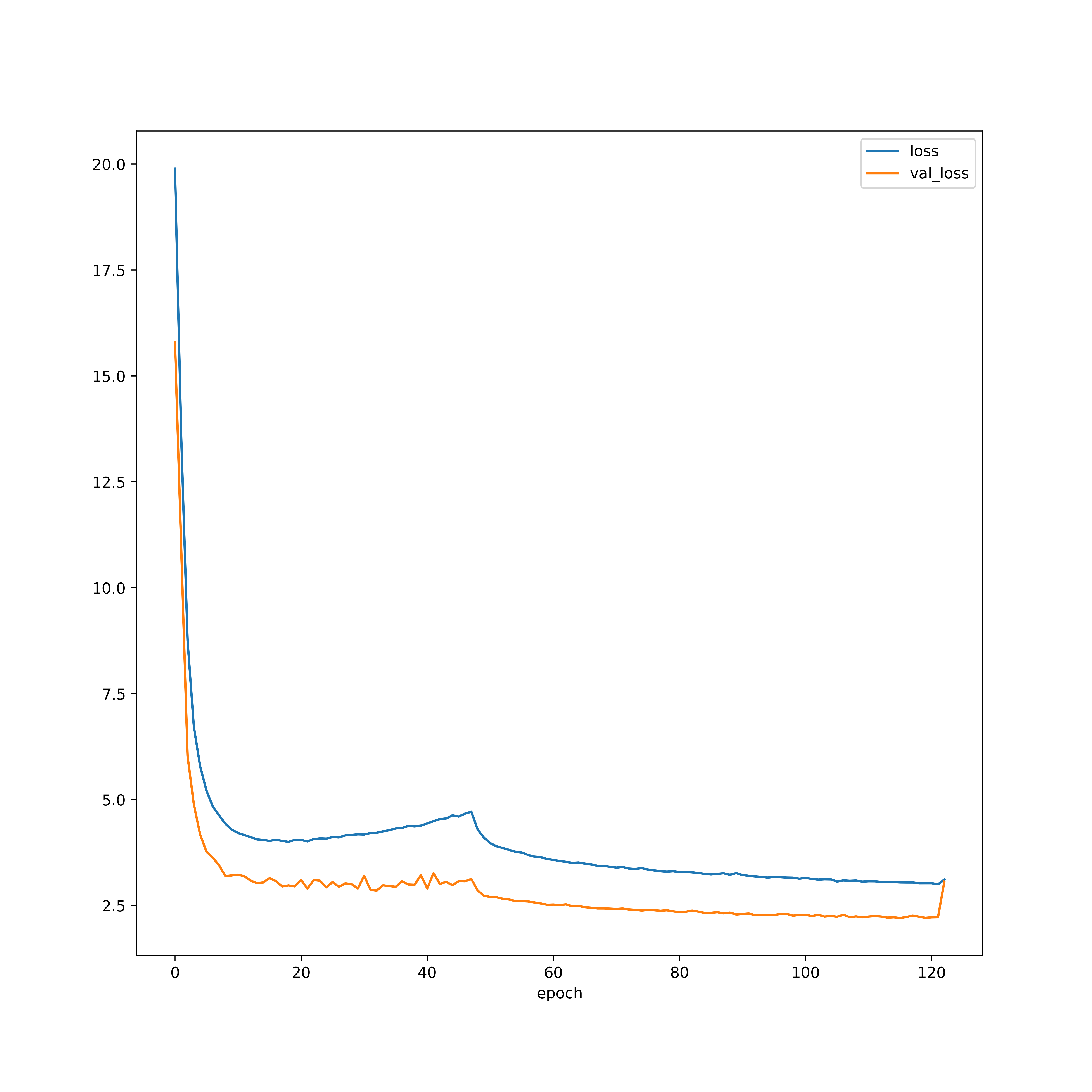}
        \caption{Puigcerver model loss}
        \label{fig:puigcerver_loss}
    \end{subfigure}
\caption{Models losses} 
\label{fig:model_losses}
\end{figure}

The first experiment was conducted with the Bluche model which showed a performance with an average of 8.36\% WER, 28.95\%  CER, and 31.45\% SER on the test dataset.  Fig. \ref{fig:bluche_loss} shows the training and validation losses of the Bluche model. 

The next experiment was carried out with the Flor model which was trained on data for 186 epochs. This model demonstrated performance with an average of 6.52\% CER, 24.52\%  WER, and 26.98\% SER on the test dataset. Fig. \ref{fig:flor_loss} shows the training and validation losses of the Flor model.

The same experiments were conducted with the Abdallah HTR model. Unlike previous models examined, this model showed a lower average of 8.22\% CER, 22.60\% WER, and 25.22\% SER on the test dataset. Fig. \ref{fig:abdo_loss} shows the training and validation losses of the Abdallah model.

The last experiment was carried out with the Puigcerver model which was trained on data for 135 epochs. This model demonstrated performance with an average of 8.01\% CER, 26.34\%  WER, and 28.94\% SER on the test dataset. Fig. \ref{fig:puigcerver_loss} shows the training and validation losses of the Puigcerver model. Table \ref{tab:Bluche_Puigcerver} shows the result of the comparison between all models.

\begin{table}[ht!]
\centering
	\caption{\footnotesize{CER, WER, and SER for Bluche, Puigcerver, Flor, and Abdallah }}
\begin{tabular}{|c|c|c|c|}
    \hline
    Model &   CER & WER & SER   \\
    \hline
    Flor  & 6.52\% & 24.52\%  &  26.98\% \\
    \hline
    Puigcerver  &  8.01\% & 26.34\% &28.94\% \\
    \hline
    Abdallah &  8.22\% & 22.60\%  & 25.22\%   \\
   
    \hline
     Bluche & 8.36\% & 28.95\%  & 31.45\% \\
    \hline

\end{tabular}

\label{tab:Bluche_Puigcerver}
\end{table}

\section{Conclusion and future work}
\label{section:Conclusion}
In this research work, firstly, we have built the Kazakh Offline Handwritten Text Dataset (KOHTD). The dataset can serve as a basis for research in handwriting recognition. This consists of a large collection of exam papers filled by students at Satbayev University and Al-Farabi Kazakh National University. Secondly, we propose a Genetic Algorithm (GA) based on random enumeration of a parameter. By random here we mean that in order to find a solution using GA, random changes were applied to the current solutions to generate new ones. GA is based on Darwin's theory of evolution. It is a slow, gradual process that works by making small and slow changes. In addition, GA is slowly making small changes to its decisions until it gets the best solution.

Finally, this research work tried to solve a handwritten Kazakh interpretation task using well-known RNN models, such as Flor, Abdallah, Bluche, and Puigcerver HTR models. These RNN models were first quantitatively evaluated against each other to select the best-performing one. According to experiments, the Flor HTR model demonstrated the highest recognition rate overall. As experiment results proved, the Flor model demonstrated the best performance with a 6.52 \% character error rate (CER), 24.52\% word error rate (WER), and 26.98\% SER for the test dataset.

In future work, The physical look of handwriting reveals the relationship between gender and handwriting. So we will present information about the gender to use for the classification of gender based on handwriting and writer identification.

\section{Acknowledgement}
We would like to thank the following people for helping with this research project:
Svetlana Sannikova,
Assel Akzhalova,
Ilyas Tursunov,
Ruslan Barmashyov,
Rassul Tolegenov,
Daniyar Kurmankhojayev,
Maksat Kanatov,
Kuanysh Slyamkhan,
Zhanibek Kaimuldenov,
Nurmukhamed Ubaidullayev,
Ardak Shalkarbay-ulyi,
Raiymbet Tukpetov, 
and
Zhuldyz-zhan Sagimbayev.
This work was funded by the Ministry of Education and Science of the Republic of Kazakhstan (Grant No AP05135175)

 \bibliographystyle{elsarticle-num} 
 \bibliography{cas-refs}





\end{document}